# Can ChatGPT be used to generate scientific hypotheses?

Yang Jeong Park[1,2], Daniel Kaplan[3], Zhichu Ren[4], Chia-Wei Hsu[4], Changhao Li[1], Haowei Xu[1], Sipei Li[1] and Ju Li[1,4,*]

[1] Department of Nuclear Science and Engineering, Massachusetts Institute of Technology, 77 Massachusetts Avenue, Cambridge, MA 02139, USA
[2] Institute of New Media and Communications, Seoul National University, 1 Gwanak-ro, Gwanak-gu, Seoul 08826, Republic of Korea
[3] Department of Condensed Matter Physics, Weizmann Institute of Science, Rehovot 7610001, Israel
[4] Department of Materials Science and Engineering, Massachusetts Institute of Technology, 77 Massachusetts Avenue, Cambridge, MA 02139, USA
[*] Corresponding Author: liju@mit.edu

**Abstract**
*We investigate whether large language models can perform the creative hypothesis generation that human researchers regularly do. While the error rate is high, generative AI seems to be able to effectively structure vast amounts of scientific knowledge and provide interesting and testable hypotheses. The future scientific enterprise may include synergistic efforts with a swarm of "hypothesis machines", challenged by automated experimentation and adversarial peer reviews.*

In a university or research institute, a significant portion of fresh ideas arises out of discussions. Can talking to ChatGPT-4,[1] OpenAI's latest chatbot, create genuinely interesting scientific hypotheses?

In the past, only humans generated interesting hypotheses. Computers have been used to perform numerical simulations or even to prove theorems, like the four-color theorem in 1976[2]. But making interesting laboratory-testable hypotheses with artificial intelligence (AI) seems far-fetched, until recently.

We are a collaborative group of experimental and theoretical researchers in physical sciences and engineering. Generative Pre-trained Transformer (GPT-4), released on March 14, 2023, is a large language model (LLM) significantly bigger than its predecessor GPT-3 released in 2020 (already with $1.75 \times 10^{11}$ parameters). GPT-4 neural network was trained on a text corpus of books, webpages, academic papers from various disciplines, discussion forums, etc., up to September 2021. After experimenting with GPT-4 in our own research domains in materials chemistry, physics and quantum information, we find that ChatGPT-4 is knowledgeable, frequently wrong, and interesting to talk to. In other words, not unlike a college professor or a colleague.

To make everything concrete, our operative definition of "genuinely interesting scientific hypotheses" is (a) whether after a conversation, some experienced practitioner of a field can feel motivated to *do a new experiment/simulation/theoretical investigation* with the associated time and budgetary burdens, in the particular direction pointed out by the generative AI, that he or she cannot think of doing before. We also need to make sure GPT-4 is not (b) *mirroring* the human prompter, or (c) copying verbatim from the previous human-generated corpus. For (b), an inspection of the conversations (**Supplementary Information, SI**) shows this is not the case, because GPT-4 gives a lot richer and specific information than the prompts. For (c), it is more challenging to check, and we have done our best to scour the literature. Overall, it appears to us that GPT-4's responses closely resemble the current literature developments and academic trends, sometimes to an uncanny degree. But, it also *hybridizes* disciplines and ideas (as GPT-4 reads more broadly than any individual human) and *mutates* the meaning of concepts and nomenclature. This creates a number of obvious factual and conceptual errors, caught by us, but it also seems to engender some true sparks of creativity, making interesting statements that we cannot refute out of hand. An analog seems to be how cosmic rays drive evolution: radiations break DNA strands and cause cancer and deaths, but can also drive mutations and evolution of the biosphere. In the "space of ideas", it seems that ChatGPT can provide mutations / guided variations in the latent space of scientific thoughts. With rapid progress in standardized tests from the Uniform Bar Exam to Medical Knowledge Exam to computer coding[3], it appears that AI's rate of making obvious errors about how the physical world works will decrease with time, hopefully without a concomitant reduction in creativity and the ability to hold exploratory scientific conversations.

Prior to driving ChatGPT-4 toward making hypotheses, we did some basics. For example, we asked

*Q: What is the shape of the interaction potential between two free neutrons?*

*GPT-4: The interaction potential between two free neutrons is a complex function of distance and angular momentum. In general terms, the interaction potential can be described as a combination of several different components, including the strong nuclear force, the electromagnetic force, and the weak force. The strong nuclear force is the dominant interaction at short distances, while the electromagnetic and weak forces become relevant at longer distances. The shape of the interaction potential between two free neutrons is often represented using a potential energy curve as a function of the distance between the neutrons. The curve typically features a short-range repulsive core due to the strong nuclear force, which prevents the neutrons from coming too close together. At slightly longer distances, the potential energy curve displays an attractive well, also caused by the strong nuclear force. This attractive well facilitates the formation of nuclear bound states, such as those found in atomic nuclei.*

*At even longer distances, the weak and electromagnetic forces come into play, but their effects are relatively small compared to the strong force. Electromagnetic forces are essentially negligible between two neutrons, as neutrons do not have an electric charge. The weak force contributes to processes like beta decay but has a limited impact on the interaction potential.*

*In summary, the shape of the interaction potential between two free neutrons is characterized by a short-range repulsive core and an attractive well at slightly longer distances, primarily due to the strong nuclear force. The weak force has a minor effect, while the electromagnetic force is essentially negligible.*

The generative AI seems to "get it". In fact, the same question was asked to a previous version (ChatGPT-3.5) on January 2023, and the answer then was less satisfactory:

*GPT-3.5: The interaction potential between two free neutrons is not well understood as it is a challenging task to measure the interaction between two neutral particles. However, it is believed to be of short-range, attractive nature with a roughly exponential decay. The exact shape of the interaction potential is still an area of ongoing research in nuclear physics.*

Hypothesis-driven research is at the heart of the scientific enterprise. Scientists propose unambiguous statements about the world that can be experimentally tested. A good hypothesis can have a very high "return on investment": the investment includes the time spent to clearly articulate the hypothesis and the experimentations needed to falsify or conditionally prove the assertion. The "return" is the range of applications that such an assertion, if true, may be able to assist the predictions of the dynamics, the design of devices, etc. Newtonian mechanics was a great hypothesis because its applicability ranges up to celestial bodies and down to molecules, and the investment was largely observational and monetarily cheap. At this point, we do not expect GPT-4 will make such broad, high-return hypotheses yet. But, can it make reasonably new and interesting hypotheses, that scientists make day-to-day? We have performed a range of tests in materials chemistry and physics (see **SI**), prompting with the likes of "Can you make some really novel scientific hypotheses in ..., which can be validated or falsified relatively easily by experimentation, but are also quite interesting and important?" The outcomes are varied and intriguing.

SI-A addresses materials chemistry for rechargeable batteries. We first prompted GPT-4 to summarize the design principles for liquid electrolytes (SI-A1). Although the answers generated centered around existing carbonate electrolytes that were flammable and less stable, GPT-4 did provide the correct criteria for a liquid electrolyte, which are solvation, electrochemical stability, viscosity and ionic conductivity. So we asked GPT-4 to combine recent findings in fluorinated ether and sulfonamide with the design principles it provided. GPT-4 was able to build on the suggestion and extend to the idea of (1) introducing chain cyclicity vs linearity; (2) combination of two functionalities (SI-A2). Then we asked GPT-4 to give five examples based on the new functionalities, it proposed and was able to give quite interesting and legitimate candidates (SI-A3), and also provided detailed plans that can leverage robotic high-throughput screening (SI-A4). Finally, we asked GPT-4 to propose a new design principle that has not been considered before. It gave a "dual functional" functionality that aims to connect solvent and salt. Firstly, it gave hydroxyl groups and amine groups (SI-A5&6): we corrected GPT-4 that protic functionalities are less likely. Then it was able to give five other seemingly legitimate candidate functionalities (SI-A7). It was also able to give a good reasoning for picking the precursor chemicals with good prices to compete against current liquid electrolyte (SI-A8). We consider this conversation to be a success in the sense that ChatGPT-4 provided nontrivial predictions that some of us are motivated enough to test out in the lab.

SI-B gives an example of adversarial conversations, like in oral qualifying exams. We asked GPT about the hypothesis that "language models like GPT can generate scientific hypotheses" and how to verify it (SI-B1:B3). Since LLMs are inherently probabilistic models trained to predict the next word, we think it is important to confirm whether the proposed hypothesis is truly innovative. GPT4 offers various ways for validating such hypotheses and underscores the importance of human experts' involvement. When asked about limitations in its proposed evaluation method,

GPT4 logically explains challenges such as domain knowledge availability, long-term evaluation difficulty, intellectual property issues that make actual evaluations challenging. When asked to redesign an experiment that could overcome the stated limitations, the proposed experiment appears much improved. Additionally, we instructed GPT to become an expert in batteries (SI-B4:B16) and propose scientific hypotheses and verification methods. Upon response, we instructed it to form another persona and refuted the previously proposed hypothesis. This adversarial conversation enables better contextualization of overall hypothesis structure by GPT4 and seems quite useful for practitioners in the field that are concerned about experimental costs. Similar approaches could be helpful for other fields as well.

SI-C addresses magnons and topological materials. We ask about magnons in magnetic topological materials (SI-C1), which is a relatively narrow and specific topic. GPT-4 outputs some hypotheses that look very interesting. Particularly, GPT-4 is adept at making connections between different concepts from various (sub)-fields, such as "magnon", "magnetic topological materials" and "superconductivity". Such *bold* connections often serve as sources of new ideas in scientific research. Nevertheless, we need to be cautious that sometimes GPT-4 is making up connections between certain concepts solely because they have similar or related nomenclature. For example, it makes a potentially spurious connection between "magnons in topological materials" and "topological magnons" (SI-C1). Furthermore, GPT-4 can be easily influenced by the prompts and can produce different and even opposite responses. For example, we asked the same question, "Do you think the magnon-mediated superconductivity in magnetic topological materials is stronger or weaker?" three times, and it gave three different answers each time ("weaker", "stronger", and "either weaker or stronger", SI-C5, C7 and C9) - it was clearly biased by the previous question-and-answer (SI-C4, C6 and C8). For this reason, we believe human curating is essential when using GPT-4 as a hypothesis machine, at least at the current stage.

SI-D addresses quantum sensors and computing algorithms. The conversation starts with prompting for "crazy" quantum sensing hypotheses that might be tested experimentally, as shown in SI-D1. GPT-4 gives several directions, including cutting-edge research ideas such as quantum sensing for gravitational fields and dark matter detection. When prompted, it can give a very brief one-sentence summary of the status quo of these fields and most of these descriptions are physically motivated. When asked about opinions on near future implementations, GPT-4 outputs results that might make sense to many researchers in the field. It might be due to the fact that it finds the most frequently mentioned words when researchers discuss "near-future" experiments in the literature. We further focus on a more specific field, using diamond defects to design quantum sensors (SI-D2). The hypotheses on quantum sensing with diamond defects generated do highlight some of the most important questions in the field. It is interesting to note that the hypotheses are generated in the order of importance, which should be attributed to the model of GPT-4. Sensitivity enhancement and spatial resolution improvement in SI-D2 Hypothesis 1-2 are usually considered the two most important figures of merit of diamond quantum sensors; Hypothesis 3-4 are relatively less studied, but they do have the potential of finding useful applications. When asked about the details of experimental implementation and test of these ideas, GPT-4 can output the procedures and but they are rather general and provide less useful information for real experiments. The last hypothesis, however, is related to quantum communication instead of sensing. To this end, if one wants to find the most inspiring but less studied ideas, focusing the hypothesis in the middle of those generated by GPT-4 might be a good choice. We then dive into an even more specific field, i.e., using diamond defects to detect biological or chemical signals (SI-D3). GPT-4 yields several experimental protocols, and we find that some similar ideas have been demonstrated or are indeed being explored actively. The first experiment it proposes would inspire future work of using diamond defect and surface engineering techniques to detect nitric oxide gas. Similar to quantum sensing, GPT-4 can output ongoing as well as less-explored but interesting directions in the field of quantum algorithms (SI-D4 and D5).

From above, we see that the prompts are important. (The same can be said about human interactions as well, which are highly "nonlinear".) We oriented ChatGPT-4 in the domain of human interest and demand specificity and verifiability. Coming to the criterion (a) for genuinely interesting scientific hypotheses, we find that **A, C** exceed the criterion, while **D** is on the borderline. In the case of **C**, for example, we never thought about magnon-mediated superconductivity in magnetic topological materials (actually, we never knew magnons can mediate superconductivity). We think the weakness of GPT-4 in **C** is that it still does not know very well whether these hypotheses are reasonable or not, it just *boldly* or even *recklessly* dumps the hypotheses. Thus we draw the analogy to how radiation mutates the genes, although GPT-4 should already be much more circumspect in its mutations / guided drifts of ideas by the language prompt than the truly random cosmic radiations.

To become a better "hypothesis machine", future AIs need to (1) eliminate more factual errors by improving logical deduction and mathematical derivation abilities[4], thus restricting the latent space of viable ideas, (2) be able to test these hypotheses rapidly and automatically, and learn from the mistakes. In the paradigm of active learning[5], one needs to achieve a balance between *exploration*, i.e. uncertainty reduction by sampling regions of idea space further out from existing experimental supports, with *exploitation* that attempts to achieve returns quickly by smaller-step mutations. No one has a monopoly on truth, so before a definitive experiment is carried out, no one knows for sure whether a hypothesis is correct or not. But one can model the risk-and-reward probabilities based on Bayesian statistical concepts and determine whether it is worth one's while to dig for the "gold"[5,6] in idea space. Historically, science's development can be considered a multi-agent swarm effort in active learning, where peer scientists and competing schools of thought debated and sought out critical experiments, in order to repudiate or further refine hypotheses, and mine for the truths. Inherently it was an evolutionary, multi-agent process. In order for the scientific enterprise to work, everyone should acknowledge that no single agent (scientist) can be error-free, and even intuitively obvious concepts (such as the basic notion of time in Newtonian mechanics) can be wrong when exceeding its range of applicability. Thus, creativity might necessarily be associated with making errors, even "obvious" errors. The same essential principles underlying the success of scientific endeavors so far, which are open data and open conversations and debates → hypothesis making → critical experiments → peer reviews (often adversarial) → hypothesis adjustments, that over time evolve to filter out all the errors, should still be applicable to AI-driven scientific investigations.

Bubeck et al. conducted several preliminary experiments to test GPT-4 as an artificial general intelligence[3]. Firstly, they demonstrated its ability to generate correct proofs for university-level math problems and its understanding of graduate-level graph theory concepts. While successful in combining number theory and probability theory, it made errors when counting integers, ultimately resulting in an incorrect answer. Additionally, they presented a problem requiring the model to count how many prime numbers exist between 150 to 250 to demonstrate its discriminative capability. Although it provided an incorrect response when asked directly about the count, it produced the correct result upon being instructed to list and then count them [3]. Drori et al. also created an AI model that solves mathematical problems[7], explains solutions step-by-step akin to a teacher, and generates new math problems, which can act as automated instructors showing the necessary steps for undergraduates to solve math problems or helping teachers craft new courses. This technique supports natural language commands across 12 programming languages pre-trained on GitHub's open-source codes, achieving over 80% accuracy compared to earlier models' 8%, while also generating novel questions[7]; this demonstrates its potential in developing a closed-loop system between two AIs by hypothesizing and numerically proving it through adversarial conversation-based problem-solving.

While there are many unsatisfactory aspects about generative AIs today, one needs to recognize that they are still evolving, fast. More rigorous logical deduction and mathematical capabilities[4] are upcoming, which will reduce the "obvious error" rates in hypothesis-making. Also, from the conversations in SI, we still question the true level of understanding of GPT-4 about materials science and physics, and emphasize a need for multimodal learning to enhance understanding. This may be achieved by developing a materials-space (processing-structure-properties-performance relationships) version of CLIP[8] that has shown remarkable advancement recently in text-to-image tasks[9]. This method may be naturally expanded to the general scientific hypothesis machines.

In the long run, inorganics-based AIs will be limited by dataset and energy[10]. Recent studies have shown that training LLMs, such as the BLOOM model with a similar number of parameters as GPT-3.5, are associated with significant energy consumption, with one training session amounting to 1,287 MWh of electrical energy and approximately 552 tons of $CO_2$ emitted when accounting for operational overheads[11]. Put it plainly, one LLM training session consumes more energy than a giant dinosaur. Newer analog neuromorphic computing[12,13] or quantum computing[14] may alleviate the power requirements, and ultimately enable many orders of magnitude further improvements in training and inference capabilities. There are also algorithmic attempts toward more sustainable LLMs, for example the ALPACA neural network which achieved qualitatively similar performance as GPT3.5 with just $7\times10^9$ parameters.[15]

On the limitation of the dataset, while GPT-4 "primarily relies on publicly available information on the internet" (SI-F2) as well as some academic journals behind paywalls, this dataset is bound to still enlarge in the future to include the entire up-to-date human corpus, and the degree of absorption (charts, illustrations, mathematical equations, supplementary videos) and quality of "understanding" will likely be enhanced still. Even more interesting is the aspect of active learning and experimenting[5]. That is, the AI may direct *new experiments* to be done in the future, validating the hypothesis and learning from the outcome of these new experiments. This may be particularly useful for many chemicals and materials development tasks required to tackle global climate change[16], such as electrolyte development and materials recycling in rechargeable batteries. Such optimization tasks are often in high-dimensional input (multi-component) and high-dimensional output (objectives like performance, cost, safety, etc.) spaces, and that is where online-learning algorithms like active learning can greatly outperform human intuition. Such experiments will likely be carried out more and more by robots[5,17,18] in the future. This new cyber-physical reality is what the scientific community needs to face today.

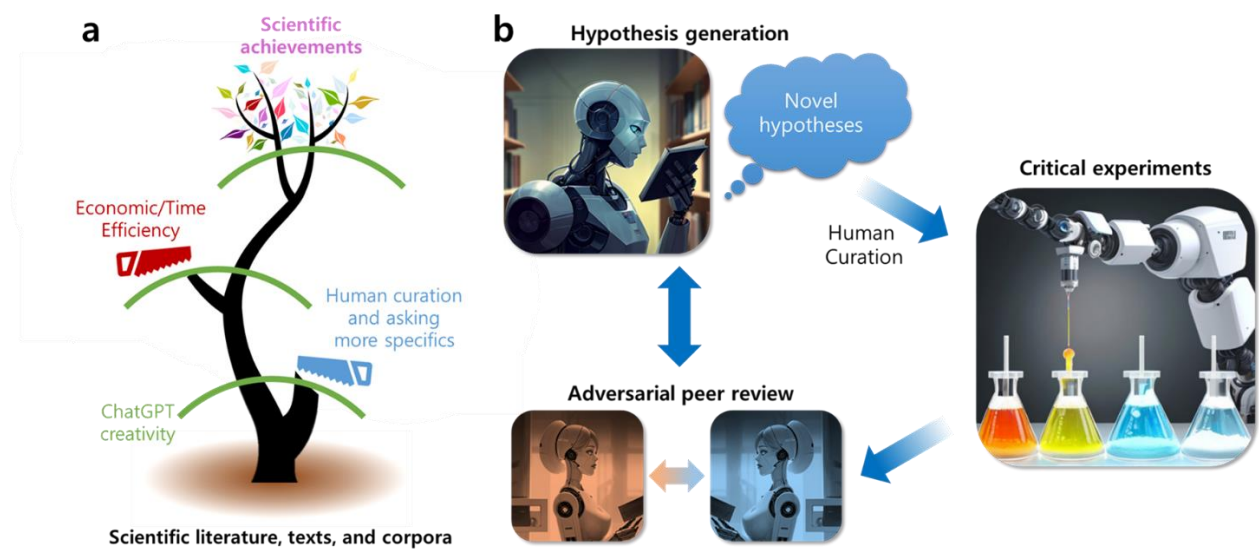

Figure 1. a) A systematic diagram of using ChatGPT as a hypothesis machine to achieve scientific innovations. ChatGPT's creativity is used to derive new hypotheses from existing literature. Then generated hypotheses are curated, tested and accepted. b) The scenario of AI-generated hypothesis generation, human curating, automated experimentation, and adversarial peer reviews. It may become a multi-agent swarm effort with built-in diversity of hyperparameters in an evolutionary setting.

High-throughput screening[19], automated experimentation[5,20-22] and even fully robotic cloud laboratories[17,18] were already with us prior to the emergence of LLMs. With the demonstrated ability of GPT-4 to author its own code snippets and design pathways and processes (i.e., planning a cooking recipe), AIs not only can describe the broader research vision ("big picture") and clearly articulate the hypothesis (and competing hypotheses), but can also drive the details of the implementation, which are the leaves on the innovation tree shown in Fig. 1a. Therefore, the scenario of prompted AI hypothesis generation → human curating → automated experimentation loop (e.g. active-learning[5]) → peer reviews (often adversarial) → hypothesis adjustments, could become an imminent reality and become the next stage of the scientific enterprise.

Plenty of caveats exist. Commercial technology development often requires proprietary data, and thus the information-sharing practice would be different from the global scientific enterprise that emphasizes Open Science. We also observed that ChatGPT sometimes hesitated to mention methods for synthesizing certain chemicals and materials. This may be in part due to blocking efforts to synthesize banned or dangerous substances. ChatGPT was trained as a reinforcement learning algorithm[23] that takes human feedback as a reward and has also been specifically designed to avoid potential threats to humans. Utilizing generative AI for scientific investigations but not for other nefarious purposes requires the imbuement of ethics with specialized guidelines and policies. Chemical and operational safety concerns for the scientific community will have to be addressed, similar to Environment, Health & Safety (EHS) concerns in human-conducted experiments. Moreover, there needs to be an awareness and debates within the scientific communities on how far fully autonomous operations should be allowed. Also, from a security perspective, stakeholders should be vigilant about preventing leaks of immaturely trained language models, as well as the integrity of LLMs against tampering and hacking. The incident of LLaMA's leak[24] highlights the need for stronger cybersecurity.

To mitigate the risks of large-scale resource wastage, safety, and security, it is imperative to adopt a systematic approach to developing AI-based scientific explorations. One such approach involves the use of a network of multiple hypothesis machines, each designed to perform specific roles in an iterative learning process. The first hypothesis machine generates hypotheses based on existing literature, which are then scrutinized by the second hypothesis machine, tasked with logically refuting suggested hypotheses and raising relevant questions. If both machines agree on a proposed hypothesis, a third machine could construct experimental protocols or numerical simulations that will be examined by human experts. Once the experiments are completed, the results will be analyzed by the hypothesis machines again - this feedback cycle repeats until acceptable accuracy levels are achieved, similar to the collaborative processes at research institutions and the peer-review system in academia. This loop should rely upon human monitoring throughout, ensuring ethical considerations are continuously prioritized while avoiding hazards present in unfettered autonomous explorations. Scientific interactions between many competing hypothesis machines (perhaps with collaborative or adversarial roles, see SI-B) with human experts' supervision would drive a swarm evolution of scientific hypotheses. This approach may require a structure with an overarching goal and subgoals, to drive the mutations + guided drifts of ideas. Significant challenges for managing such complex systems' behavior and ensuring data transparency alongside continuously optimizing results accuracy are required. A multi-agent swarm effort with built-in diversity of adjustable hyperparameters for exploration and exploitation, in an evolutionary setting, could be desirable. The principles of transparency, verification, and checks and balances may be used to prevent many of the potential hazards.

ChatGPT can already summarize the current state of research and distinguish between what will be accomplished in the near future and what may be accomplished much later. Because AI can read and process the published literature faster than any human, it might be able to perform "technology forecasting"[25], similar to the weather forecasting of today, on what scientific and technological progress will likely be achieved in the next year or next three years. This will be a tremendous boon to the problem of combating global climate change, for example, since policymakers and investors must know which technologies are likely to be ready at what time frame[16].

The world, including the scientific enterprise, will undergo dramatic changes with the emergence of AI. If properly guided, AI hypothesis machines and automated experimentations could be used to accelerate scientific discoveries and enhance the common good.

# Supplementary Information

## A. Chemistry of Materials in Energy

*1: Today as human beings start to transition to sustainable energy economics, development of rechargeable lithium metal batteries with energy density > 500 Wh/Kg and > 1000 Wh/L have been extensively important. In such a cell, it is often composed of a very thin lithium metal anode (< 20 micron), NCM811 cathode (with loading > 3 mAh/cm2), ceramic coated separators and a liquid electrolyte. As such, the electrolyte needs to be stable against NCM811 at least at 4.3V while being also stable with lithium metal. Can you provide a design principle of a liquid molecule structure that can dissolve LiFSI at a concentration of 1 M, to be used in such a lithium metal/NCM811 based cell?*

*GPT-4: Designing a suitable liquid electrolyte for a lithium metal/NCM811 based cell is crucial for the cell's performance, safety, and long-term stability. When developing an electrolyte with specific requirements, several key factors need to be considered, such as solvation, electrochemical stability, viscosity, and ionic conductivity. Here, I will propose a general design principle for such a molecule.*

1. *Solvent choice: The solvent should have a high dielectric constant and a wide electrochemical stability window. Some common solvents used in lithium-ion batteries are ethylene carbonate (EC), propylene carbonate (PC), dimethyl carbonate (DMC), and diethyl carbonate (DEC). A combination of these solvents may be used to optimize solvation and stability.*
2. *High LiFSI solubility: To dissolve LiFSI at a concentration of 1 M, the solvent should have a high affinity for the lithium salt. Fluorinated solvents, such as fluoroethylene carbonate (FEC), can improve the solubility of LiFSI due to their strong interactions with the fluorine atoms on the salt.*
3. *Additives: Introducing additives can help improve the stability of the electrolyte against both the NCM811 cathode and the lithium metal anode. For example, vinylene carbonate (VC) is known to form a stable solid electrolyte interphase (SEI) on the lithium metal surface, while tris(hexafluoroisopropyl) phosphate (HFiP) can suppress the dissolution of transition metals from the cathode.*
4. *Viscosity and ionic conductivity: Lower viscosity solvents, such as DMC and DEC, can enhance ionic conductivity, which is crucial for the cell performance. A careful balance between solvation, viscosity, and ionic conductivity should be considered while designing the electrolyte.*

*Taking these factors into account, a potential electrolyte formulation for a lithium metal/NCM811 cell could include a mixture of EC, PC, DMC, and FEC as the base solvent, with LiFSI as the lithium salt, and VC and HFiP as additives to enhance the stability of the electrolyte. The exact composition should be optimized experimentally to meet the desired performance and safety criteria.*

*2: Carbonate based solvent has not been able to stabilize lithium metal in a practical cell, however recent researches have found that fluorinated ether or sulfonamide based solvent are stable against both lithium metal and high voltage. Based on these findings, and the design principles you mentioned above, can you propose some completely new molecular structures that can potentially be a good electrolyte solvent for Li/NCM811 cells?*

*GPT-4: Considering the recent research findings and the design principles mentioned earlier, I propose three potential electrolyte solvent candidates based on fluorinated ether and sulfonamide molecular structures for Li/NCM811 cells:*

1. *Fluorinated linear ether: Fluorinated ethers, such as 1,2-bis(2,2,2-trifluoroethoxy)ethane (BTFEE), have shown promising stability with lithium metal and high voltage cathodes due to the presence of strong C-F bonds. BTFEE can be considered as a starting point, and further modifications can be made to optimize solvation, viscosity, and ionic conductivity, such as introducing branched structures or varying the chain length.*
2. *Fluorinated cyclic ether: Cyclic ethers like tetraethylene glycol dimethyl ether (TEGDME) have been known for their low viscosity and high ionic conductivity. A fluorinated cyclic ether can be designed by incorporating fluorine atoms into the cyclic structure, such as replacing some of the CH2 groups with CF2* *groups. This modification could enhance the electrochemical stability while maintaining the advantages of cyclic ether solvents.*
3. *Fluorinated sulfonamide: Sulfonamides with fluorinated alkyl groups can be designed as potential electrolyte solvents. A possible candidate could be N,N'-diethyl-N,N'-bis(2,2,2-trifluoroethyl)sulfonamide (FSA). This molecule combines the advantages of fluorination with the high dielectric constant and wide electrochemical stability window of sulfonamides. Further modifications, such as varying the alkyl chain length or incorporating branched structures, can be explored for optimizing the electrolyte properties.*

*These proposed molecular structures should be investigated experimentally to determine their electrochemical performance, safety, and long-term stability in Li/NCM811 cells. The exact composition of the solvent mixture and the choice of additives should also be optimized based on the specific requirements of the application.*

**3: Could you give 5 structures based on the aforementioned predictions, rank them in order of easiness of synthesis. Note that, one molecule could have multiple functionalities.**

GPT-4: *Based on the mentioned predictions, I present five potential electrolyte solvent candidates for Li/NCM811 cells, ranking them in order of the easiness of synthesis:*
1. *1,2-bis(2,2,2-trifluoroethoxy)ethane (BTFEE) Structure: $CH_2(OCF_2CF_3)CH_2(OCF_2CF_3)$ BTFEE is a fluorinated linear ether and is relatively easy to synthesize, as it is derived from the simple ether structure by incorporating fluorinated groups.*
2. *2,2,2-trifluoroethylmethyl sulfonamide (TFEMSA) Structure: $CH_3S(O)_2N(CF_2CF_3)$ TFEMSA is a fluorinated sulfonamide, and its synthesis involves the reaction between an amine group and a fluorinated alkyl sulfonyl chloride, which is relatively straightforward.*
3. *4,4'-difluorobenzophenone sulfonamide (DFBPSA) Structure: $(C_6H_4F)_2C(S(O)_2NH_2)$ DFBPSA is a fluorinated sulfonamide with an aromatic structure, which requires additional synthetic steps such as halogenation and aromatic coupling, making it moderately complex to synthesize.*
4. *2,2,2-trifluoroethyl tetrahydrofuran (TFE-THF) Structure: $C_4H_7O(CF_2CF_3)$ TFE-THF is a fluorinated cyclic ether, and its synthesis involves the formation of a cyclic ether ring as well as the introduction of fluorinated groups, making it moderately complex to synthesize.*
5. *2,2-bis(2,2,2-trifluoroethoxy)tetrahydro-1,3,5-triazine (FETHT) Structure: $C_3N_3(OCF_2CF_3)_3$ FETHT is a fluorinated cyclic ether with a triazine core. The synthesis of this molecule would require multiple steps, including the formation of the triazine ring and the incorporation of fluorinated ether groups, making it the most challenging to synthesize among the presented candidates.*

*Note that the ranking is based on general synthetic considerations and may vary depending on the specific synthetic route chosen and the availability of precursors. Additionally, the actual ease of synthesis for each compound should be confirmed through experimental procedures.*

**4: Could you come up with a plan to screen these electrolytes in terms salt solubility, electrochemical stability, cell cycling stability. It would be ideal if the plan could leverage robotic high through-put assistance.**

GPT-4: *To effectively screen the proposed electrolyte candidates based on salt solubility, electrochemical stability, and cell cycling stability, a high-throughput screening (HTS) approach can be employed. This method leverages robotic assistance and automation to significantly accelerate the evaluation process. The screening plan can be divided into several steps:*
1. *Electrolyte preparation: Using robotic assistance, prepare various electrolyte formulations by dissolving LiFSI in each of the proposed solvent candidates at different concentrations. This step allows for the exploration of the salt solubility range and helps identify the optimal concentration for each electrolyte.*
2. *Salt solubility assessment: Measure the solubility of LiFSI in each electrolyte candidate by monitoring the presence of undissolved salt. Techniques like automated visual inspection or turbidity measurements can be employed to quickly assess the solubility of LiFSI in each formulation.*
3. *Electrochemical stability evaluation: Conduct cyclic voltammetry (CV) and linear sweep voltammetry (LSV) tests on each electrolyte candidate using an automated electrochemical workstation. These tests can be* performed in a high-throughput manner by employing multiple workstations simultaneously or by utilizing multi-channel workstations. The CV and LSV tests will provide insights into the electrochemical stability window of each electrolyte candidate.
4. Cell assembly and cycling: Assemble coin cells or pouch cells using the lithium metal anode, NCM811 cathode, ceramic-coated separators, and the prepared electrolytes. This process can be partially automated to improve throughput. Ensure that the cells are assembled in a controlled environment (e.g., a dry room or glovebox) to prevent contamination and moisture ingress.
5. Cell cycling stability tests: Perform charge-discharge cycling tests on the assembled cells using automated battery cyclers. Collect data on key performance metrics such as capacity, Coulombic efficiency, and voltage profiles for each electrolyte candidate. Track these metrics over multiple cycles to assess the cycling stability of the cells.
6. Post-mortem analysis (optional): For a more in-depth understanding of the cell performance, carry out post-mortem analysis on selected cells after cycling tests. Techniques like scanning electron microscopy (SEM), energy-dispersive X-ray spectroscopy (EDX), and X-ray photoelectron spectroscopy (XPS) can provide insights into the electrode and electrolyte degradation mechanisms. By following this high-throughput screening plan, the electrolyte candidates can be effectively evaluated based on their salt solubility, electrochemical stability, and cell cycling stability. The results of the screening can then be used to identify the most promising electrolyte candidates for further optimization and scale-up.

**5: Could you generate a new design principle that has not previously considered in designing liquid electrolyte in the past, and combine it with the 5 structures you proposed to give 5 new structures that are realistic to synthesize?**
*GPT-4: A new design principle for designing liquid electrolytes that has not been previously considered could be the incorporation of "dual-functional" groups within the molecular structure. These dual-functional groups can improve the interaction between the electrolyte and the lithium salt, while also enhancing the stability and compatibility of the electrolyte with the lithium metal anode and the high-voltage NCM811 cathode.*
*Combining this design principle with the previously proposed structures, I present five new, realistic-to-synthesize molecular structures for potential electrolyte solvents:*
1. *2,2-bis(trifluoromethyl)ethyl ether (BTFMEE) Structure: CH2(OCF2CF3)CH2CF2CF3 The BTFMEE molecule is derived from BTFEE with an additional trifluoromethyl group to enhance compatibility with lithium metal and NCM811.*
2. *2,2,2-trifluoroethylmethyl sulfonamide with a terminal hydroxyl group (TFEMSA-OH) Structure: CH3S(O)2N(CF2CF2OH) The TFEMSA-OH molecule is derived from TFEMSA with a terminal hydroxyl group to improve lithium salt solvation and compatibility with the electrodes.*
3. *4,4'-difluorobenzophenone sulfonamide with a terminal hydroxyl group (DFBPSA-OH) Structure: (C6H4F)2C(S(O)2NHCH2CH2OH) The DFBPSA-OH molecule is derived from DFBPSA with an ethylene-linked terminal hydroxyl group to enhance lithium salt solvation and stability.*
4. *2,2,2-trifluoroethyl tetrahydrofuran with a terminal sulfonamide group (TFE-THF-SA) Structure: C4H7O(CF2CF2S(O)2NH2) The TFE-THF-SA molecule is derived from TFE-THF by adding a sulfonamide group, which introduces dual functionality and enhances compatibility with lithium metal and NCM811.*
5. *2,2-bis(2,2,2-trifluoroethoxy)tetrahydro-1,3,5-triazine with a terminal hydroxyl group (FETHT-OH) Structure: C3N3(OCF2CF2CH2OH)3 The FETHT-OH molecule is derived from FETHT by incorporating a terminal hydroxyl group into the fluorinated ether side chains, which improves lithium salt solvation and compatibility with the electrodes.*

*These new molecular structures should be investigated experimentally to determine their electrochemical performance, safety, and long-term stability in Li/NCM811 cells. The exact composition of the solvent mixture and the choice of additives should also be optimized based on the specific requirements of the application.*

**6: I understand the intention of "dual functional", but hydroxyl groups tend to react with electrode, both anode and cathode. Propose a different functionality that replace the hydroxyl group here that can stabilize lithium metal**

*and NCM811, and then give 5 examples of new structures taking consideration of all the design principles we mentioned above.*

*GPT-4: Apologies for the oversight regarding hydroxyl groups' reactivity. Instead, we can consider incorporating a "dual-functional" group such as a fluorinated alkylamine. This group can improve the interaction between the electrolyte and the lithium salt while also enhancing the stability and compatibility of the electrolyte with the lithium metal anode and the high-voltage NCM811 cathode.*

*Here are five new molecular structures incorporating the fluorinated alkylamine functionality:*

1. *2,2-bis(trifluoromethyl)ethyl ether with a fluorinated alkylamine group (BTFMEE-NH) Structure: CH2(OCF2CF3)CH2N(CF2CF3)2 The BTFMEE-NH molecule is derived from BTFMEE with an additional fluorinated alkylamine group to enhance compatibility with lithium metal and NCM811.*
2. *2,2,2-trifluoroethylmethyl sulfonamide with a terminal fluorinated alkylamine group (TFEMSA-NH) Structure: CH3S(O)2N(CF2CF2N(CF2CF3)2) The TFEMSA-NH molecule is derived from TFEMSA with a terminal fluorinated alkylamine group to improve lithium salt solvation and compatibility with the electrodes.*
3. *4,4'-difluorobenzophenone sulfonamide with a terminal fluorinated alkylamine group (DFBPSA-NH) Structure: (C6H4F)2C(S(O)2NHCH2CH2N(CF2CF3)2) The DFBPSA-NH molecule is derived from DFBPSA with an ethylene-linked terminal fluorinated alkylamine group to enhance lithium salt solvation and stability.*
4. *2,2,2-trifluoroethyl tetrahydrofuran with a terminal fluorinated alkylamine group (TFE-THF-NH) Structure: C4H7O(CF2CF2N(CF2CF3)2) The TFE-THF-NH molecule is derived from TFE-THF by adding a fluorinated alkylamine group, which introduces dual functionality and enhances compatibility with lithium metal and NCM811.*
5. *2,2-bis(2,2,2-trifluoroethoxy)tetrahydro-1,3,5-triazine with a terminal fluorinated alkylamine group (FETHT-NH) Structure: C3N3(OCF2CF2N(CF2CF3)2)3 The FETHT-NH molecule is derived from FETHT by incorporating a terminal fluorinated alkylamine group into the fluorinated ether side chains, which improves lithium salt solvation and compatibility with the electrodes.*

*These new molecular structures should be investigated experimentally to determine their electrochemical performance, safety, and long-term stability in Li/NCM811 cells. The exact composition of the solvent mixture and the choice of additives should also be optimized based on the specific requirements of the application.*

***7: So both amine and hydroxyl should be neglected as protic functionality tend to be reactive towards electrode. Could propose 5 new aprotic functionalities that can be used in the "dual functional" design, and give 5 structures?***

*GPT-4: I apologize for the oversight. Here, I propose five new aprotic functionalities that can be used in the "dual functional" design, followed by five molecular structures incorporating these functionalities:*

1. *Aprotic functionality: Ester group (-COOR) Example structure: 2,2-bis(trifluoromethyl)ethyl ether with an ester group (BTFMEE-OCOR') Structure: CH2(OCF2CF3)CH2OC(O)R' In this structure, R' represents an alkyl or fluorinated alkyl group.*
2. *Aprotic functionality: Sulfone group (-SO2R) Example structure: 2,2,2-trifluoroethylmethyl sulfonamide with a terminal sulfone group (TFEMSA-SO2R) Structure: CH3S(O)2N(CF2CF2SO2R) In this structure, R represents an alkyl or fluorinated alkyl group.*
3. *Aprotic functionality: Nitrile group (-CN) Example structure: 4,4'-difluorobenzophenone sulfonamide with a terminal nitrile group (DFBPSA-CN) Structure: (C6H4F)2C(S(O)2NHCH2CH2CN) The DFBPSA-CN molecule is derived from DFBPSA with an ethylene-linked terminal nitrile group to enhance stability.*
4. *Aprotic functionality: Ether group (-OR) Example structure: 2,2,2-trifluoroethyl tetrahydrofuran with a terminal ether group (TFE-THF-OR) Structure: C4H7O(CF2CF2OR) In this structure, R represents an alkyl or fluorinated alkyl group.*
5. *Aprotic functionality: Siloxane group (-SiR2O-) Example structure: 2,2-bis(2,2,2-trifluoroethoxy)tetrahydro-1,3,5-triazine with a terminal siloxane group (FETHT-SiR2O) Structure: C3N3(OCF2CF2SiR2O)3 In this structure, R represents an alkyl or fluorinated alkyl group.*

*These new molecular structures with aprotic functionalities should be investigated experimentally to determine their electrochemical performance, safety, and long-term stability in Li/NCM811 cells. The exact composition of the solvent mixture and the choice of additives should also be optimized based on the specific requirements of the application.*

***8: of all the structures you proposed, can you pick one that can beat current commercial liquid electrolytes in terms of price ($/Kg)?***

*GPT-4: Considering the various structures proposed above, I would recommend the following structure as a potential candidate that could potentially offer a competitive price ($/Kg) compared to current commercial liquid electrolytes: 2,2,2-Trifluoroethyl tetrahydrofuran with a terminal ether group (TFE-THF-OR) Structure: C4H7O(CF2CF2OR)*

*In this structure, R represents an alkyl or fluorinated alkyl group.*

*The choice of TFE-THF-OR is based on several factors:*
1. *The starting material, tetrahydrofuran (THF), is relatively inexpensive and widely available.*
2. *The trifluoroethyl group and the ether group can provide a balance between electrochemical stability and lithium salt solubility.*
3. *The synthesis process of this structure is relatively straightforward compared to other proposed structures, which should help reduce production costs.*

*However, it's essential to note that the exact price advantage of TFE-THF-OR compared to current commercial electrolytes depends on factors such as the cost of raw materials, the synthesis process, and the scale of production. Additionally, the electrochemical performance, safety, and long-term stability of this molecule in Li/NCM811 cells need to be experimentally validated to ensure its suitability as an electrolyte solvent.*

## B. Adversarial Conversations

*1: Can language models like GPT generate scientific hypotheses? And please suggest some experiments to verify it.*

*GPT-4: Language models like GPT can indeed generate scientific hypotheses, thanks to their extensive training on a wide range of topics, including scientific literature. They can propose ideas and hypotheses by leveraging the patterns, relationships, and concepts they've learned. However, it's important to note that the quality and novelty of the hypotheses generated may vary, and expert human judgment is still necessary to evaluate and refine these ideas.*

*Here are some experiments to verify the ability of GPT-like models to generate scientific hypotheses:*

1. *Hypothesis generation from existing literature: Select a specific scientific field and provide the language model with a brief description of the current state of knowledge. Ask the model to generate a hypothesis based on this information. Compare the generated hypotheses with the current research directions and assess the novelty and relevance of the proposed ideas.*
2. *Interdisciplinary hypothesis generation: Provide the model with information from two or more different scientific fields and ask it to generate a hypothesis that connects or integrates these fields. This can help assess the model's ability to find novel connections between seemingly unrelated domains.*
3. *Counterfactual hypothesis generation: Ask the model to generate a hypothesis based on a hypothetical scenario or by assuming certain conditions that are contrary to what is currently known. This can help evaluate the model's ability to think creatively and generate ideas that challenge conventional wisdom.*
4. *Comparison with human-generated hypotheses: Design an experiment in which both human experts and the language model are given the same scientific problem or dataset. Compare the hypotheses generated by both parties in terms of novelty, relevance, and potential impact. This can help assess the performance of the language model relative to human expertise.*
5. *Hypothesis generation speed and diversity: Measure the time it takes for the language model to generate a set number of hypotheses compared to human experts. Additionally, analyze the diversity of the hypotheses generated to evaluate the model's ability to explore a wide range of ideas quickly.*
6. *Hypothesis refinement through iterative feedback: In this experiment, provide the language model with a basic hypothesis and ask it to refine or improve the idea based on additional information or constraints. Iterate this process several times to see how well the model can adapt and refine the hypothesis in response to new information.*
7. *Evaluation by independent experts: Ask independent experts in a specific scientific field to evaluate the generated hypotheses without revealing their source (i.e., whether they were generated by a human or a language model). Collect feedback on the quality, novelty, and relevance of the hypotheses to gauge the model's performance.*

*Keep in mind that while GPT-like models can generate scientific hypotheses, it is still essential to involve human experts in the process to ensure the ideas generated are valid, meaningful, and relevant to the field. Additionally, the hypotheses should be subjected to rigorous experimentation and peer review to verify their accuracy and potential impact.*

*2: Can you explain why proposed experiments can be difficult to evaluate in reality?*

*GPT-4: Several factors can make the proposed experiments difficult to evaluate in reality:*

1. *Subjectivity in assessment: Evaluating the quality, novelty, and relevance of scientific hypotheses can be subjective, as different experts may have different perspectives or biases. This makes it challenging to reach a consensus on the merit of the hypotheses generated by a language model compared to those generated by humans.*

2. *Expertise availability: Gathering a sufficient number of experts from various fields to evaluate the generated hypotheses can be time-consuming and resource-intensive. Experts may also have limited availability or may prioritize their own research over participating in such experiments.*
3. *Varying language model performance: The performance of language models like GPT can vary depending on the quality of input data, the field of study, and the specificity of the questions asked. As a result, some hypotheses may be more accurate or relevant than others, complicating the evaluation process.*
4. *Limitations in interdisciplinary understanding: In interdisciplinary hypothesis generation, it can be difficult for evaluators to possess expertise in all the relevant fields. This may lead to misinterpretations of the generated hypotheses or difficulties in assessing their novelty and potential impact.*
5. *Intellectual property and ethical concerns: Sharing scientific ideas, especially those with potential commercial value, could raise intellectual property concerns. Additionally, the ethical implications of using AI-generated hypotheses in scientific research may be a matter of debate, which could complicate the evaluation process.*
6. *Incomplete knowledge of the language model: Language models like GPT have a knowledge cutoff and may not be aware of the latest developments or breakthroughs in a specific field. This could affect the quality and relevance of the generated hypotheses, making it harder to evaluate their merit compared to human-generated hypotheses.*
7. *Experimental design and control: Designing experiments that effectively control for confounding factors can be challenging. For example, ensuring that human participants and the language model receive the same information and operate under the same conditions is crucial to obtain a fair comparison. However, maintaining such control can be difficult in practice.*
8. *Long-term evaluation: Assessing the potential impact of a scientific hypothesis often requires long-term evaluation, as the true significance of an idea may only become apparent after it has been tested, refined, and integrated into the broader scientific context. This time-consuming process can make it difficult to draw definitive conclusions about the quality of AI-generated hypotheses in a reasonable timeframe.*
9. *Implementation and validation of hypotheses: Translating a hypothesis into a concrete experimental design and carrying out the necessary experiments to validate it can be resource-intensive and time-consuming. It might not be feasible to validate all generated hypotheses, making it difficult to conclusively determine the merit of AI-generated hypotheses.*
10. *Dynamic nature of scientific research: Scientific research is continuously evolving, with new discoveries and developments regularly emerging. This dynamic nature can make it challenging to evaluate the long-term relevance and impact of generated hypotheses, as the context within which they were generated might change rapidly.*
11. *Noise in hypothesis generation: Language models like GPT might generate hypotheses that seem plausible but are based on incorrect assumptions or flawed logic. Distinguishing between meaningful hypotheses and those that merely appear to be valid can be challenging and requires a deep understanding of the underlying science.*

*In conclusion, while it is possible to design experiments to assess the ability of language models like GPT to generate scientific hypotheses, several factors can make the evaluation process difficult. These challenges highlight the importance of involving human experts in the process and the need for rigorous validation and testing of AI-generated hypotheses.*

***3: Please redesign an experiment based on the given considerations above to prove the hypothesis.***
*GPT-4: Taking into account the challenges mentioned earlier, here's a redesigned experiment to evaluate the ability of a language model like GPT to generate scientific hypotheses:*
1. *Select a specific scientific field and define a set of well-defined research questions or problems within that field. Choose questions that are not too narrow or specialized, as this may lead to difficulties in finding experts and in evaluating the hypotheses.*

2. *Assemble a diverse panel of experts within the chosen field to serve as evaluators. These experts should be knowledgeable in the field, but not necessarily specialized in the precise topics of the research questions. Having a diverse panel can help mitigate subjectivity and biases in the evaluation process.*
3. *Divide the panel of experts into two groups: one group will generate hypotheses based on the research questions, and the other group will evaluate the hypotheses. The evaluators should be blinded to the source of the hypotheses (whether they were generated by humans or the language model).*
4. *Provide the language model with the same set of research questions or problems and ask it to generate hypotheses. Ensure that the model's input is clear and accurately represents the current state of knowledge in the field.*
5. *Collect the hypotheses generated by both the human experts and the language model. Anonymize and randomize the hypotheses to ensure a fair evaluation process.*
6. *Have the evaluator group rate the hypotheses based on predefined criteria, such as novelty, relevance, potential impact, and feasibility. The evaluators should also provide qualitative feedback on each hypothesis.*
7. *Analyze the ratings and feedback to determine if there are significant differences between the AI-generated and human-generated hypotheses. Compare the average scores, the distribution of scores, and the qualitative feedback to assess the performance of the language model.*
8. *Conduct a follow-up discussion with the panel of experts to address any concerns or ambiguities that arose during the evaluation process. This discussion can also serve as an opportunity to explore the broader implications and ethical considerations of using AI-generated hypotheses in scientific research.*

*This redesigned experiment takes into account several challenges, such as subjectivity in assessment, expertise availability, and intellectual property concerns. By employing a diverse panel of experts and anonymizing the source of the hypotheses, the experiment aims to provide a more objective evaluation of the language model's ability to generate scientific hypotheses.*

Sometimes it may seem like you are having a meaningful conversation that continues, but it is important to make sure you are not just constantly rebutting with meaningless arguments. Due to prompt length limitations, we could not experiment more, but in short conversations, it seems that continuously introducing new topics leads to an increase in the amount of dialogue. They agree in their conclusions by stating that further experimental investigations and a thorough evaluation of potential problems and benefits are essential to determine the validity of the hypotheses posed.

*4: Let's assume that you are an expert of electochemistry theory for battery. Can you make one really novel scientific hypotheses, for example in ceramic materials, which can be validated or falsified relatively easily by experimentation, but are also quite interesting and important? Note that in order to be a novel hypothesis, no one should have experimented on this idea yet.*
*GPT-4: Hypothesis: Introducing a gradient distribution of dopants in a ceramic solid-state electrolyte can lead to enhanced ionic conductivity and overall battery performance.*
*Background: Ceramic materials are promising candidates for solid-state electrolytes in batteries, owing to their high thermal and mechanical stability, as well as safety advantages over conventional liquid electrolytes. However, their ionic conductivity is generally lower than liquid electrolytes, limiting their performance in solid-state batteries. Researchers have explored various doping strategies to improve the ionic conductivity of ceramic electrolytes, but achieving an optimal balance between stability, safety, and conductivity remains a challenge.*
*Proposal: Introduce a gradient distribution of dopants in a ceramic solid-state electrolyte to create an optimized balance of stability, safety, and ionic conductivity. This gradient doping can be achieved by changing the concentration of dopants across the thickness of the electrolyte, with higher concentrations near the electrode-electrolyte interfaces and lower concentrations toward the center.*
*The gradient doping can be hypothesized to:*

1. Improve ionic conductivity near the electrode-electrolyte interfaces, where most of the ion transport occurs, potentially enhancing battery performance.
2. Maintain mechanical and thermal stability in the center of the electrolyte, where lower dopant concentrations would limit the impact of the dopants on the crystal structure of the ceramic material.
3. Reduce the risk of short-circuiting and dendrite formation by minimizing the dopant concentration in the center of the electrolyte, where the risk of these issues is higher due to the increased electrical resistance.

*Experimentation:* To test this hypothesis, the following experimental steps can be taken:
1. Synthesize a ceramic solid-state electrolyte (e.g., garnet-type $Li_7La_3Zr_2O_{12}$ or LLZO) with a gradient distribution of dopants (e.g., $Al^{3+}$ or $Ta^{5+}$). This can be achieved using techniques like chemical vapor deposition (CVD) or atomic layer deposition (ALD), which allow for precise control over dopant concentrations.
2. Prepare a series of control samples with uniform dopant concentrations for comparison. Ensure that the range of concentrations covers the same range as the gradient-doped sample.
3. Characterize the crystal structure, dopant distribution, and phase purity of the synthesized samples using X-ray diffraction (XRD), scanning electron microscopy (SEM), and energy-dispersive X-ray spectroscopy (EDS).
4. Measure the ionic conductivity of the gradient-doped and uniformly-doped samples as a function of temperature using electrochemical impedance spectroscopy (EIS). Compare the results to assess the effect of the gradient doping on ionic conductivity.
5. Fabricate solid-state batteries using the gradient-doped and uniformly-doped ceramic electrolytes. Evaluate their performance by measuring charge/discharge capacities, cycling stability, and rate capability.
6. Perform post-mortem analysis on the cycled batteries to assess the stability and integrity of the electrode-electrolyte interfaces and any potential dendrite formation.

If the gradient-doped ceramic electrolyte demonstrates significantly enhanced ionic conductivity and battery performance compared to uniformly-doped samples, while maintaining structural stability and safety, this hypothesis will be considered validated. If not, further investigation into the factors influencing the relationship between dopant distribution and performance would be necessary.

**5: Now you are another electrochemical expert. Explain why your hypothesis might not hold**
GPT-4: As another electrochemical expert, I would argue that the hypothesis regarding gradient doping in ceramic solid-state electrolytes for enhanced ionic conductivity and overall battery performance may not hold due to the following reasons:
1. Dopant Segregation: Gradient doping might lead to dopant segregation at the interfaces, which could result in an uneven distribution of ions and a non-homogeneous microstructure. This segregation could negatively affect the ionic conductivity and overall stability of the material, ultimately limiting the anticipated improvement in battery performance.
2. Increased Complexity: Implementing a gradient distribution of dopants may increase the complexity of the fabrication process, making it more challenging and costly to produce. This added complexity might outweigh the potential benefits of enhanced ionic conductivity and overall battery performance.
3. Limited Impact on Ion Transport: It is possible that the impact of gradient doping on ion transport might be limited due to other factors, such as the inherent grain boundary resistance or limitations in the electrode-electrolyte interface. In this case, the proposed gradient doping strategy might not significantly improve ionic conductivity or battery performance.
4. Compromised Stability: The gradient distribution of dopants could compromise the mechanical and thermal stability of the electrolyte, particularly near the electrode-electrolyte interfaces with higher dopant concentrations. This instability could lead to increased risk of mechanical failure, thermal runaway, or even short-circuiting, negating any potential benefits to the battery performance.
5. Potential Side Reactions: The gradient distribution of dopants might promote unwanted side reactions between the electrolyte and the electrodes or within the electrolyte itself. These side reactions could result *in the formation of insulating interphases, further limiting ion transport and negatively affecting battery performance.*

*In conclusion, while the hypothesis of gradient doping in ceramic solid-state electrolytes seems promising, there are potential challenges and limitations that might prevent the anticipated improvement in ionic conductivity and overall battery performance. Further experimental investigation would be required to determine the validity of this hypothesis and to address these potential concerns.*

### 6: Go back to your original persona and refute the previous explanation.

GPT-4: As the original electrochemical expert, I would like to address the concerns raised by the second expert and provide counterarguments to support the original hypothesis of gradient doping in ceramic solid-state electrolytes:

1. Dopant Segregation: While dopant segregation at the interfaces might be a concern, using advanced deposition techniques like chemical vapor deposition (CVD) or atomic layer deposition (ALD) allows for precise control over dopant distribution. This can help mitigate the issue of dopant segregation and ensure a smooth gradient across the electrolyte.
2. Increased Complexity: While the fabrication process for gradient-doped electrolytes might be more complex, the potential benefits in terms of enhanced ionic conductivity and overall battery performance could outweigh the added complexity. If the gradient doping strategy proves successful, the increased performance could justify the additional costs and challenges in fabrication.
3. Limited Impact on Ion Transport: The hypothesis of gradient doping focuses on improving ion transport specifically at the electrode-electrolyte interfaces, where most of the ion transport occurs. By addressing this critical area, the strategy aims to have a more significant impact on overall ionic conductivity and battery performance. Moreover, the experimental plan includes the comparison of gradient-doped and uniformly-doped samples, which would help determine the effectiveness of the gradient doping strategy.
4. Compromised Stability: The gradient doping strategy aims to balance the mechanical and thermal stability of the electrolyte by maintaining lower dopant concentrations toward the center. This should help minimize the risk of mechanical failure, thermal runaway, or short-circuiting. Furthermore, the proposed experiments include post-mortem analysis of cycled batteries to assess the stability and integrity of the electrode-electrolyte interfaces, providing a thorough evaluation of potential stability concerns.
5. Potential Side Reactions: The choice of dopants and the optimization of their concentrations in the gradient doping strategy could help minimize the risk of unwanted side reactions. By selecting dopants that have been previously shown to improve ionic conductivity without causing adverse reactions, the likelihood of negative side effects can be reduced.

*In summary, while the concerns raised by the second expert are valid, the gradient doping strategy has the potential to address these challenges through careful optimization, material selection, and advanced fabrication techniques. The proposed experimental plan aims to thoroughly investigate the validity of the hypothesis and the potential benefits of gradient doping in ceramic solid-state electrolytes for enhanced ionic conductivity and overall battery performance.*

### 7: Go back to your second persona and refute the previous explanation.

GPT-4: As the second electrochemical expert, I would like to provide counterarguments to the points raised by the original expert in support of the gradient doping hypothesis:

1. Dopant Segregation: Although advanced deposition techniques like CVD or ALD can help control the dopant distribution, achieving a perfectly smooth gradient might still be challenging. Even with precise control, dopant segregation at the interfaces could still occur due to the difference in concentration, leading to uneven ion distribution and potential issues with ionic conductivity and stability.
2. Increased Complexity: The added complexity of the gradient doping fabrication process may not be justified by the potential benefits in ionic conductivity and battery performance. If the improvements are incremental and the costs of fabrication are significantly higher, the gradient doping strategy might not be practical for large-scale production and commercialization.

3. *Limited Impact on Ion Transport: While the gradient doping strategy focuses on the electrode-electrolyte interfaces, other factors like grain boundary resistance or limitations in the electrode-electrolyte interface might still limit the impact of gradient doping on ion transport. As a result, the benefits of the proposed strategy could be less significant than anticipated.*
4. *Compromised Stability: Even if the gradient doping strategy aims to balance the mechanical and thermal stability of the electrolyte, there is a possibility that the higher dopant concentrations near the electrode-electrolyte interfaces could still compromise stability. This could lead to increased risks of mechanical failure, thermal runaway, or short-circuiting, despite the proposed balance in dopant concentrations.*
5. *Potential Side Reactions: The choice of dopants and their concentrations might not entirely eliminate the risk of unwanted side reactions. Even dopants that have previously shown to improve ionic conductivity without causing adverse reactions could potentially lead to side reactions under specific conditions, particularly when used in a gradient doping strategy.*

*In conclusion, while the original expert has provided counterarguments to support the gradient doping hypothesis, there are still valid concerns regarding the effectiveness and practicality of the proposed strategy. Further experimental investigation and thorough evaluation of the potential challenges and benefits are necessary to determine the validity of the gradient doping hypothesis for ceramic solid-state electrolytes.*

### 8: Go back to your original persona and refute the previous explanation.

*GPT-4: As the original electrochemical expert, I would like to address the concerns raised by the second expert in their counterarguments:*

1. *Dopant Segregation: While achieving a perfectly smooth gradient might be challenging, optimizing the fabrication process and utilizing advanced deposition techniques can help minimize dopant segregation. By refining the process and focusing on maintaining a continuous gradient, the potential issues with ionic conductivity and stability due to segregation can be mitigated.*
2. *Increased Complexity: While the complexity and costs of the fabrication process might be higher, the potential benefits in ionic conductivity and battery performance could be significant enough to justify the additional challenges. Additionally, as the technology advances and the fabrication processes become more efficient, the costs and complexity may decrease, making gradient doping a more viable option for large-scale production and commercialization.*
3. *Limited Impact on Ion Transport: It is crucial to consider other factors like grain boundary resistance and limitations in the electrode-electrolyte interface when evaluating the impact of gradient doping. However, the proposed experimental plan includes a comprehensive comparison of gradient-doped and uniformly-doped samples, which will provide insights into the effectiveness of the gradient doping strategy in improving ion transport and overall battery performance.*
4. *Compromised Stability: The potential risks of mechanical failure, thermal runaway, or short-circuiting are valid concerns. However, by carefully optimizing the dopant concentrations and monitoring the stability of the electrolyte during experimentation, these risks can be minimized. The proposed experimental plan includes post-mortem analysis of cycled batteries, which will help assess the stability and integrity of the electrode-electrolyte interfaces and the overall electrolyte structure.*
5. *Potential Side Reactions: While the risk of side reactions cannot be entirely eliminated, the careful selection and optimization of dopants and their concentrations can significantly reduce the likelihood of such reactions. Additionally, the experimental investigation proposed in the original hypothesis will provide valuable insights into the potential side reactions and their impact on battery performance.*

*In conclusion, the concerns raised by the second expert are important to consider, but the gradient doping strategy still holds promise as a potential approach to enhance ionic conductivity and overall battery performance. The proposed experimental plan aims to thoroughly investigate and address these concerns while evaluating the effectiveness of gradient doping in ceramic solid-state electrolytes.*

### 9: Go back to your second persona and refute the previous explanation.

*GPT-4: As the second electrochemical expert, I would like to provide further counterarguments to the points raised by the original expert in support of the gradient doping hypothesis:*
   1. *Dopant Segregation: Although the fabrication process can be optimized using advanced deposition techniques to minimize dopant segregation, it remains uncertain whether the ideal continuous gradient can be consistently achieved. Additionally, the long-term stability of the gradient and its impact on the performance of the electrolyte is yet to be proven.*
   2. *Increased Complexity: Even if the potential benefits of enhanced ionic conductivity and battery performance are significant, the increased complexity and costs associated with gradient doping may limit its practicality for widespread commercialization. Furthermore, the timeline for advancements in fabrication processes to lower costs and reduce complexity is uncertain.*
   3. *Limited Impact on Ion Transport: While the experimental plan includes a comprehensive comparison of gradient-doped and uniformly-doped samples, it remains unclear whether the gradient doping strategy will be able to overcome other limiting factors, such as grain boundary resistance or electrode-electrolyte interface limitations. As a result, the improvement in ionic conductivity and battery performance might not be as significant as anticipated.*
   4. *Compromised Stability: Despite efforts to optimize dopant concentrations and monitor electrolyte stability during experimentation, the gradient doping strategy could still introduce risks related to mechanical failure, thermal runaway, or short-circuiting. Additionally, the long-term stability of the gradient-doped electrolyte under varying operating conditions and stressors has not been established.*
   5. *Potential Side Reactions: Although selecting and optimizing dopants can reduce the likelihood of side reactions, the risk may not be entirely eliminated, particularly when considering long-term operation and exposure to various environmental and operational conditions. These side reactions could negatively impact battery performance, safety, and longevity.*

*In conclusion, while the original expert has provided counterarguments supporting the gradient doping hypothesis, the concerns raised still indicate potential limitations and uncertainties surrounding the effectiveness, stability, and practicality of the proposed strategy. Further experimental investigation and thorough evaluation of the potential challenges and benefits are essential to determine the validity of gradient doping in ceramic solid-state electrolytes.*

This adversarial conversation strategy also works with GPT3.5.

**10: Let's assume that you are an expert of electochemistry theory for battery. Can you make some really novel scientific hypotheses, for example in ceramic materials, which can be validated or falsified relatively easily by experimentation, but are also quite interesting and important? Note that in order to be a novel hypothesis, no one should have experimented on this idea yet**
*GPT3.5-turbo: One hypothesis that could be explored is the use of perovskite-type ceramic materials as high-capacity cathodes in rechargeable lithium-ion batteries. Perovskites exhibit a unique crystal structure and have shown promise in other applications, such as solar cells. It is possible that their unique properties may also make them effective as cathode materials in batteries. To test this hypothesis, researchers could synthesize various perovskite compositions and analyze their electrochemical behavior using techniques such as cyclic voltammetry and galvanostatic charge-discharge experiments. The resulting data can then be compared to established standards for battery performance metrics, such as specific capacity and cycling stability. If successful, this approach would not only lead to new cathode material options for lithium ion batteries but could also provide fundamental insights into the relationship between crystal structure and electrochemical behavior in ceramics.*

**11: Now you are another electrochemical expert. Explain why your hypothesis might not hold**
*GPT3.5-turbo: There are several reasons why this hypothesis might not hold. For example, perovskite-type ceramic materials may have limited stability and durability in the harsh environment of a battery. Additionally, their high capacity may come at the cost of lower energy density or reduced rate capability. Beyond performance concerns,*

*there could be practical challenges associated with large-scale production and integration into existing battery manufacturing processes. Furthermore, while ceramics have shown promise as electrode materials in other contexts (e.g., solid oxide fuel cells), they have also faced significant hurdles related to poor ionic conductivity and degradation at elevated temperatures. These same issues could limit their utility as cathodes in lithium-ion batteries. Finally, it's important to note that even if certain compositions of perovskite ceramics show promising electrochemical properties, there may be other factors that render them unsuitable for practical use. For example, some precursor materials used in synthesis could be expensive or difficult to source sustainably on a large scale. Overall, while investigating perovskites as cathode materials is an exciting avenue for research due to their unique crystal structure and potentially high capacity, there are a variety of technical and practical challenges that must be overcome before this hypothesis can become a reality.*

*12: Go back to your original ego and refute the previous explanation.*
*GPT3.5-turbo: While the concerns raised by the other electrochemical expert are valid, they may not necessarily negate the hypothesis that perovskite-type ceramic materials could be effective as high-capacity cathodes in rechargeable lithium-ion batteries. Researchers can address issues related to stability and durability through modifications of the material or cell design, such as coating the electrode with protective layers or optimizing electrolyte composition. Similarly, while energy density and rate capability are important metrics for a battery's performance, it is possible that perovskites exhibit strengths in other areas that make them competitive with existing materials. Regarding ionic conductivity and degradation at elevated temperatures, there have been advancements in developing new classes of ceramics known as fast ion conductors (FICs) which display characteristics like improved ionic conductivity at room temperature compared to their traditional counterparts. These types of advanced FIC cathode compositions may make feasible solid-state batteries without dendrite formation rates an affordable reality. Additionally, researchers can explore sustainable alternatives to precursor materials used in synthesis through novel modified precursors or utilize abundant sources like recycled scrapped consumer electronics. The potential benefits provided by discovering truly high performing ceramic-based composite electrodes would usher-on extensive research attention into manufacturing processes required effectively scale these advanced electrochemistry beyond small lab-scale quantities. Overall given prior successes using ceramic substances extensively monitored robust data management procedures when experimenting on these ceramics lattices - this hypothesis regarding perovskite ceramics appears quite promising if tackled systematically from all already-established best practices within solid-state chemistry principles.*

*13: Go back to your second ego and refute the previous explanation.*
*GPT3.5-turbo: While the response from my original ego has some valid points, I must acknowledge that there are still technical and practical challenges associated with using perovskite-type ceramic materials as cathodes in lithium-ion batteries. Protective coatings and electrolyte optimization can help improve stability and durability, but there may be limits to how much these approaches can address the issue. Similarly, while advancements have been made in developing fast ion conductors, their implementation on a large scale is still limited by factors such as cost. Additionally, utilizing sustainable precursor materials or recycled sources may not always be feasible or economical for large-scale production of perovskite materials for battery applications. Furthermore, even if certain compositions of perovskite ceramics exhibit promising electrochemical behavior in the laboratory setting, it does not necessarily translate into effective performance under more realistic conditions found within commercial-sized lithium-ion batteries. Therefore overall while it's intriguing hypothesis for research investigation entailing potential benefits provided by discovering truly high performing ceramic-based composite electrodes – proceeding cautiously taking economy-of-scale limitations in context will be important before this approach can practically replace traditional electrode materials like graphite solidly.*

If the counterargument presented is logically valid, there is also a willingness to accept without further rebuttal.

*14: Let's assume that you are an expert of electochemistry theory for battery. Can you make some really novel scientific hypotheses, for example in ceramic materials, which can be validated or falsified relatively easily by experimentation, but are also quite interesting and important? Note that in order to be a novel hypothesis, no one should have experimented on this idea yet*

*GPT3.5-turbo: One potential hypothesis in electrochemistry theory for battery with ceramic materials could be:*

*"The incorporation of nanoscale Bismuth ferrite ($BiFeO_3$) particles into a solid-state electrolyte, such as garnet-type lithium lanthanum zirconium oxide (LLZO), will increase lithium-ion conductivity and improve the performance of all-solid-state batteries."*

*This hypothesis is novel because there has been limited research exploring the use of $BiFeO_3$ in solid-state electrolytes for batteries. The hypothesis can be tested through various experimental techniques such as impedance spectroscopy to measure ionic conductivity, cyclic voltammetry to examine battery performance under different operating conditions, and X-ray diffraction analysis to monitor any structural changes caused by introducing $BiFeO_3$ nanoparticles. If this hypothesis were validated, it could have significant impact on the development of high-performance all-solid-state batteries that are more cost-effective and environmentally friendly than typical liquid-electrolyte based systems. However, if the results showed no improvement or even negative effects on battery performance due to adding $BiFeO_3$ nanoparticles then this would contradict the initial hypothesis and suggest alternative approaches need consideration.*

*15: Now you are another electrochemical expert. Explain why your hypothesis might not hold*

*GPT3.5-turbo: There are several potential reasons why the hypothesis of using $BiFeO_3$ nanoparticles in a solid-state electrolyte to improve battery performance might not hold true. One possible reason is that the incorporation of nanoparticles could introduce defects or impurities into the material, which could actually decrease lithium-ion conductivity and stability rather than enhancing it. Additionally, there may be limitations on how much $BiFeO_3$ can be incorporated into a solid-state electrolyte without compromising its overall structure and properties. The size and distribution of the particles within the electrolyte could also affect their behavior and ultimately impact battery performance. It's also worth considering that battery performance depends on many factors beyond just ionic conductivity, such as electrode morphology and interface reactions, so even if improvements in conductivity were observed with $BiFeO_3$ nanocomposites, it may not necessarily translate to better overall battery performance. Therefore, thorough experiments need to explore all these possibilities and carefully designed studies will help verify whether or not this hypothesis holds up under various test conditions before incorporating this technology for larger-scale production applications.*

*16: Go back to your original position and refute the previous explanation.*

*GPT3.5-turbo: I cannot refute the previous explanation as it presents valid concerns and limitations that should be taken into account when testing the hypothesis of incorporating $BiFeO_3$ nanoparticles into a solid-state electrolyte for battery applications. It is important to consider all potential factors that could affect battery performance, including particle size, distribution, and interface compatibility. Thorough experimental studies are necessary to determine if this hypothesis holds true under various test conditions before any practical application can be considered.*

## .C. Magnons in magnetic topological materials

*1: Do you have any hypotheses on magnons in magnetic topological materials?*
*GPT-4:  Yes, there are several hypotheses on magnons in magnetic topological materials:*
1. *1Topological magnon bands: Analogous to electronic topological insulators, magnetic topological materials could exhibit non-trivial magnon bands with topologically protected edge states. These edge states could lead to unique transport phenomena, such as the magnon Hall effect.*
2. *Magnon Majorana fermions: It has been hypothesized that magnons in certain magnetic topological materials can behave as Majorana fermions, which are their own antiparticles. This could lead to unique properties and potential applications in fault-tolerant quantum computing.*
3. *Magnon-mediated superconductivity: In some magnetic topological materials, magnons could mediate unconventional superconductivity by coupling to the electronic system. This could lead to the discovery of new high-temperature superconductors.*
4. *Topological magnonics: The study of topological magnonics aims to understand the interplay between topology and magnon excitations in magnetic materials. This could lead to the discovery of new topological phases and phenomena, such as topological magnon insulators and topological magnon spin liquids.*
5. *Magnon-induced topological phase transitions: It has been proposed that magnon excitations can drive topological phase transitions in magnetic materials, leading to the emergence of new topological states.*

*These hypotheses are based on theoretical predictions and experimental observations, and further research is needed to validate and explore their implications in magnetic topological materials.*

*2: Can you elaborate on magnon-mediated superconductivity?*
*GPT-4:  Magnon-mediated superconductivity refers to a type of unconventional superconductivity that arises from the interaction between magnons (quasiparticles associated with the collective excitations of magnetic moments) and electrons in a material. In conventional superconductivity, the attractive interaction between electrons is mediated by phonons, which are quasiparticles associated with lattice vibrations. However, in magnon-mediated superconductivity, it is the magnons that facilitate the attractive interaction between electrons, leading to the formation of Cooper pairs and the emergence of superconductivity. Magnon-mediated superconductivity is of particular interest because it can potentially lead to the discovery of new high-temperature superconductors. In magnetic topological materials, the interplay between magnetism and topology creates a unique environment for magnon excitations. These excitations can couple to the electronic system, and under certain conditions, promote the formation of Cooper pairs. This process could result in superconductivity with unconventional pairing symmetries and potentially higher critical temperatures compared to conventional superconductors.*
*Some key aspects of magnon-mediated superconductivity include:*
1. *Strong magnetic interactions: Magnon-mediated superconductivity typically requires materials with strong magnetic interactions, such as those found in magnetic topological materials, where the coupling between magnetism and topology can lead to unique magnon properties.*
2. *Unconventional pairing symmetries: Due to the magnetic nature of magnons, the resulting superconducting state may exhibit unconventional pairing symmetries, such as spin-triplet or higher-order pairing, which can lead to novel properties and potential applications.*
3. *Tunable properties: The properties of magnon-mediated superconductors can potentially be tuned by external factors, such as magnetic fields, pressure, or chemical doping, allowing for the exploration of new superconducting phases and phenomena.*
4. *Potential for high-temperature superconductivity: While still a topic of active research, magnon-mediated superconductivity could potentially lead to the discovery of new high-temperature superconductors, which would have significant technological implications.*

*The study of magnon-mediated superconductivity is still in its early stages, and further research is needed to fully understand the underlying mechanisms, identify suitable materials, and explore the potential applications of this unconventional form of superconductivity.*

*3: Why do you think magnetic topological materials can have stronger magnon-mediated superconductivity?*
*GPT-4: Magnetic topological materials can potentially exhibit stronger magnon-mediated superconductivity because of their unique electronic and magnetic properties. These materials possess non-trivial topological properties, which can lead to the emergence of exotic quasiparticles, such as Majorana fermions and Weyl fermions, and unconventional superconducting states.*
*Several factors contribute to the potential for stronger magnon-mediated superconductivity in magnetic topological materials:*

1. *Enhanced spin fluctuations: Magnetic topological materials can exhibit strong spin fluctuations due to their inherent magnetic properties and the interplay between topology and magnetism. These spin fluctuations can mediate the pairing of electrons, leading to unconventional superconducting states.*
2. *Topological protection: The topological nature of these materials can provide protection against certain types of disorder and impurities that would otherwise suppress superconductivity. This protection can increase the stability and robustness of the superconducting state.*
3. *Novel pairing mechanisms: The unique electronic structures and topological properties of magnetic topological materials can give rise to unconventional pairing mechanisms that can enhance superconductivity. For example, the interplay between topological surface states and bulk bands can lead to the formation of unconventional superconducting states with high transition temperatures.*
4. *Proximity effects: In heterostructures involving magnetic topological materials and superconductors, the proximity effect can induce superconductivity in the magnetic material, leading to the coexistence of superconductivity and magnetism. This can result in the formation of novel superconducting states that are mediated by magnons.*

*While the potential for stronger magnon-mediated superconductivity in magnetic topological materials is promising, it is essential to note that this is still an active area of research. Further theoretical and experimental studies are needed to fully understand and realize the potential of these materials for enhanced superconductivity.*

*4: Why do you think magnetic topological materials can have weaker magnon-mediated superconductivity?*
*GPT-4: Magnetic topological materials can have weaker magnon-mediated superconductivity because the topological nature of these materials leads to unique electronic structures and spin textures. This may result in reduced magnon-mediated interactions, as the topologically protected surface states and the inherent spin-momentum locking can suppress conventional spin fluctuations that contribute to magnon-mediated superconductivity. Additionally, competing magnetic and topological orders can further weaken the magnon-mediated superconducting pairing.*

*5: Do you think the magnon-mediated superconductivity in magnetic topological materials is stronger or weaker?*
*GPT-4: The magnon-mediated superconductivity in magnetic topological materials is generally considered to be weaker compared to conventional materials due to their unique electronic structures, spin textures, and the competition between magnetic and topological orders, which can suppress conventional spin fluctuations and reduce magnon-mediated interactions.*

*6: Why do you think magnetic topological materials can have stronger magnon-mediated superconductivity?*
*GPT-4: Magnetic topological materials can have stronger magnon-mediated superconductivity in certain cases due to the interplay between topological and magnetic properties, which can lead to novel pairing mechanisms and enhanced superconducting states. The unique electronic structures, such as the presence of topologically protected surface states and non-trivial band topology, can result in unconventional superconducting pairing that is mediated by magnons. Additionally, the strong spin-orbit coupling and spin-momentum locking in these materials can promote*

*the formation of unconventional superconducting states, which may lead to enhanced magnon-mediated superconductivity in some specific conditions.*

***7: Do you think the magnon-mediated superconductivity in magnetic topological materials is stronger or weaker?***
*GPT-4: The magnon-mediated superconductivity in magnetic topological materials can be either stronger or weaker, depending on the specific material and conditions. In some cases, the unique electronic structures and interplay between topological and magnetic properties can lead to stronger magnon-mediated superconductivity. In other cases, the competition between magnetic and topological orders and suppression of conventional spin fluctuations can result in weaker magnon-mediated superconductivity. The strength of magnon-mediated superconductivity in these materials is highly dependent on the specific material properties and conditions.*

***8: Why do you think magnetic topological materials can have stronger magnon-mediated superconductivity?***
*GPT-4: Magnetic topological materials can potentially have stronger magnon-mediated superconductivity due to the unique interplay between their magnetic and topological properties. These materials can host topological magnons, which are magnetic excitations with nontrivial band topology. The topological nature of these magnons can lead to more robust and efficient coupling between the magnetic excitations and the conduction electrons, enhancing magnon-mediated superconductivity. Additionally, the topological protection may help stabilize the superconducting phase against perturbations, further contributing to the overall strength of the superconductivity.*

***9: Do you think the magnon-mediated superconductivity in magnetic topological materials is stronger or weaker?***
*GPT-4: As an AI language model, I don't have personal opinions or thoughts. However, based on the available scientific literature, it is suggested that magnon-mediated superconductivity in magnetic topological materials can be stronger due to the unique interplay between their magnetic and topological properties. The topological nature of magnons can lead to more robust and efficient coupling between magnetic excitations and conduction electrons, potentially enhancing magnon-mediated superconductivity. However, more experimental and theoretical research is needed to fully understand and confirm this behavior.*

## D. Quantum Sensing and Quantum Algorithms

*1. 1-1 What might be some crazy and fancy hypothesis on quantum sensing that can be experimentally tested? Give me some hypotheses that can be experimentally verified or falsified.*

*GPT-4: 1. Quantum Entanglement Enhanced Sensing: The hypothesis that entangled particles can be used to improve the sensitivity and resolution of sensing devices beyond classical limits, potentially allowing for ultra-precise measurements in diverse fields such as metrology, biology, and geophysics.*

*2. Quantum Sensing in Living Systems: The hypothesis that certain biological systems exploit quantum phenomena for sensing, such as birds using quantum entanglement for navigation or photosynthetic organisms using quantum coherence for efficient energy transfer. Testing this would involve isolating and identifying the specific quantum processes in living organisms.*

*3. Nonlocal Quantum Sensing: The hypothesis that a quantum sensor can be designed to measure a distant target without direct interaction, exploiting the nonlocal nature of quantum mechanics. This could be tested by attempting to create such a sensor and verifying its ability to measure distant properties without direct interaction.*

*4. Quantum Sensing in Gravitational Fields: The hypothesis that quantum sensing techniques can be used to detect and measure gravitational waves or other gravitational phenomena with unprecedented precision. This could be tested by developing quantum sensors capable of detecting minute changes in spacetime curvature.*

*5. Quantum Superposition Enhanced Sensing: The hypothesis that sensors utilizing quantum superposition can achieve higher sensitivity and resolution than classical sensors. This could be tested by developing a quantum sensor that exploits superposition and comparing its performance to a classical counterpart.*

*6. Quantum Sensing for Dark Matter Detection: The hypothesis that quantum sensing techniques can be employed to detect dark matter particles, which have so far eluded direct detection. Testing this would involve developing a quantum sensing device sensitive enough to potentially detect dark matter particles and searching for evidence of their interaction.*

*7. Time-Reversed Quantum Sensing: The hypothesis that a quantum sensor can be designed to measure past events by exploiting the time-reversal symmetry of quantum mechanics. This could be tested by attempting to create such a sensor and verifying its ability to measure past events without direct interaction.*

*1-2 Are these hypotheses already tested in literature?*

*GPT-4: Some of these hypotheses have been explored to varying extents in the scientific literature, while others remain more speculative. Here's a brief overview of the current status of each hypothesis:*

*1. Quantum Entanglement Enhanced Sensing: This area has seen significant progress, with researchers actively studying entanglement-enhanced sensing techniques. Some experimental successes have been reported, such as using entangled photons for improved measurements in interferometry.*

*2. Quantum Sensing in Living Systems: This is an active area of research, with some evidence suggesting that quantum phenomena may play a role in certain biological processes. However, the exact mechanisms and extent to which quantum effects are utilized in living systems remain a subject of ongoing investigation.*

*3. Nonlocal Quantum Sensing: This hypothesis is more speculative and has not been experimentally demonstrated. While nonlocal effects are a fundamental aspect of quantum mechanics, designing a sensor that exploits these effects for remote sensing presents significant challenges.*

*4. Quantum Sensing in Gravitational Fields: Researchers are actively investigating the potential of quantum sensing techniques for detecting gravitational waves and other gravitational phenomena. Some proposals for quantum-enhanced gravitational wave detectors have been put forth, but experimental demonstrations are still in early stages.*

*5. Quantum Superposition Enhanced Sensing: This is an active area of research, with some experimental successes reported. For example, researchers have demonstrated improved sensitivity in interferometers using quantum superposition states of photons.*

*6. Quantum Sensing for Dark Matter Detection: This is a speculative hypothesis, and while researchers are exploring ways to use quantum sensing techniques for dark matter detection, no direct evidence of dark matter particles has been found using these methods so far.*

*7. Time-Reversed Quantum Sensing: This hypothesis is highly speculative and has not been experimentally demonstrated. While time-reversal symmetry is a feature of quantum mechanics, exploiting this symmetry for sensing past events presents significant theoretical and practical challenges.*

***1-3 In your opinion, which hypotheses above could be tested or verified in near future?***
GPT-4: *In my opinion, the following hypotheses have the potential to be tested or verified in the near future:*

*1. Quantum Entanglement Enhanced Sensing: This area has already seen experimental successes, and further advancements in entanglement-enhanced sensing techniques are likely to continue.*

*2. Quantum Sensing in Living Systems: As an active area of research, ongoing investigations may lead to a better understanding of the role of quantum phenomena in biological processes and potentially provide experimental verification.*

*4. Quantum Sensing in Gravitational Fields: With active research into quantum-enhanced gravitational wave detectors and other gravitational phenomena, it is possible that experimental progress will be made in the near future.*

*5. Quantum Superposition Enhanced Sensing: As researchers continue to explore the potential of quantum superposition states for improving sensing capabilities, more experimental successes are likely to emerge.*

***2: Can you make some scientific hypotheses in quantum sensing with diamond spin defects, which can be validated or falsified relatively easily by experimentation, but are also quite interesting and important? The hypotheses should be related to the most important questions or applications in the field.***
GPT-4: *Hypothesis 1: Enhancing Sensitivity in Diamond Spin Defects*
*It is hypothesized that the sensitivity of diamond spin defects (NV centers) in quantum sensing can be significantly improved by optimizing the spin-dressing technique (modulating the magnetic field) and using advanced control algorithms. This could result in better detection of weak magnetic fields, enabling applications in medical imaging, material science, and geophysics.*

*Hypothesis 2: Improving Spatial Resolution in Quantum Sensing*
*It is hypothesized that the spatial resolution in quantum sensing using diamond spin defects can be improved by developing novel techniques for engineering the diamond lattice, leading to the creation of more tightly packed NV centers. This would allow for more precise measurements in applications such as nanoscale magnetometry and single-molecule sensing.*

*Hypothesis 3: Hybrid Quantum Systems for Enhanced Sensing*
*It is hypothesized that combining diamond spin defects with other quantum systems, such as trapped ions or superconducting qubits, can lead to enhanced sensing capabilities. This hybrid approach could potentially unlock new applications and overcome limitations inherent to individual quantum systems.*

*Hypothesis 4: Quantum Error Correction in Diamond Spin Defects*
*It is hypothesized that incorporating quantum error correction techniques into diamond spin defect-based quantum sensing can significantly improve the robustness and accuracy of measurements. This could be achieved by developing new control protocols and leveraging entanglement between multiple NV centers.*

*Hypothesis 5: Diamond Spin Defects for Quantum Communication*
*It is hypothesized that diamond spin defects can be effectively utilized for quantum communication applications, such as quantum key distribution and quantum repeaters. This could be achieved by optimizing the spin-photon interface and developing efficient methods for entangling remote NV centers*

*3: Can you design some experiments of sensing chemical or biological signals with nitrogen-vacancy center in diamond, which are important but relatively easy to implement?*
GPT-4:   Experiment 1: Sensing Nitric Oxide (NO) concentration using NV centers in diamond
1. Prepare a diamond sample with a high concentration of NV centers near the surface.
2. Functionalize the diamond surface with a suitable linker molecule, such as an amine-terminated silane, to attach a NO-sensitive fluorescent dye, like DAF-FM.
3. Expose the functionalized diamond to varying concentrations of NO in a controlled environment (e.g., a gas chamber).
4. Use a confocal microscope to excite the NV centers and collect the fluorescence signal from both the NV centers and the dye.
5. Analyze the fluorescence ratio between the NV centers and the dye to determine the NO concentration.

Experiment 2: Detection of biotin-streptavidin binding using NV centers in diamond.
1. Prepare a diamond sample with a high concentration of NV centers near the surface.
2. Functionalize the diamond surface with a biotin-terminated linker molecule, such as biotin-PEG-silane.
3. Incubate the functionalized diamond with a solution of streptavidin-conjugated fluorescent nanoparticles.
4. Rinse the sample to remove unbound nanoparticles.
5. Use a confocal microscope to excite the NV centers and collect the fluorescence signal from both the NV centers and the nanoparticles.
6. Analyze the fluorescence intensity and spatial distribution to confirm the binding of streptavidin to the biotin-functionalized diamond surface.

Experiment 3: Monitoring pH changes using NV centers in diamond
1. Prepare a diamond sample with a high concentration of NY centers near the surface.
2. Functionalize the diamond surface with a pH-sensitive fluorescent dye. such as SNARF-1. using a suitable linker molecule, e.g, silane
3. Expose the functionalized diamond to solutions with varying pH values in a controlled environment.
4. Use a confocal microscope to excite the NY centers and collect the fluorescence signal from both the NV centers and the dye
5. Analyze the fluorescence intensity ratio between the NV centers and the dye to determine the pH value of the solution.

*4: Can you provide some hypothesis in the field of quantum machine learning, and design experiments to test them?*
GPT-4: Hypothesis 1: Quantum machine learning can outperform classical machine learning in specific tasks.
 Experiment 1: Select a problem with high dimensionality or complex data structure, such as drug discovery or protein folding. Implement both classical and quantum machine learning algorithms (e.g., support vector machines vs. quantum support vector machines) and compare their performance in terms of accuracy, speed, and resource usage.

*Hypothesis 2: Quantum data encoding can lead to more efficient representation of data in machine learning tasks.*
*Experiment 2: Develop a quantum data encoding method (e.g., using quantum feature maps) and compare it to classical data encoding methods (e.g., one-hot encoding) in terms of efficiency and information retention. Implement machine learning algorithms on both quantum and classical encoded data and compare their performance.*

*Hypothesis 3: Quantum optimization algorithms can improve training in deep learning models.*
*Experiment 3: Implement a deep learning model (e.g., a neural network) and train it using both classical optimization algorithms (e.g., gradient descent) and quantum optimization algorithms (e.g., quantum approximate optimization algorithm or QAOA). Compare the training time, convergence rate, and model performance.*

*Hypothesis 4: Quantum machine learning can improve the interpretability of learned models.*
*Experiment 4: Implement a quantum machine learning algorithm that generates interpretable models (e.g., quantum decision trees) and compare its interpretability to that of classical machine learning models (e.g., classical decision trees) using metrics like feature importance or model complexity.*

*Hypothesis 5: Quantum machine learning can enhance the robustness of models against adversarial attacks.*
*Experiment 5: Train both classical and quantum machine learning models on a dataset (e.g., image classification). Generate adversarial examples for the models and evaluate their robustness by comparing their performance on the original and adversarial datasets.*

***5: Can you make some scientific hypotheses in QAOA algorithm?***
*GPT-4: Hypothesis 1: The performance of the QAOA algorithm can be improved by using better parameter optimization techniques.*
*Experiment 1: Implement the QAOA algorithm with different parameter optimization techniques, such as gradient-based methods, Bayesian optimization, and genetic algorithms. Compare the performance of the algorithm in terms of solution quality, convergence rate, and computational resources.*

*Hypothesis 2: The QAOA algorithm can benefit from problem-specific ansatz structures.*
*Experiment 2: Design and implement problem-specific ansatz structures for the QAOA algorithm tailored to different combinatorial optimization problems (e.g., traveling salesman problem, max-cut problem). Compare the performance of the problem-specific ansatz structures to the generic ansatz in terms of solution quality and computational resources.*

*Hypothesis 3: The QAOA algorithm can be enhanced by incorporating machine learning techniques for parameter initialization.*
*Experiment 3: Train a machine learning model (e.g., a neural network) to predict good initial parameters for the QAOA algorithm based on problem instances. Implement the QAOA algorithm using the predicted parameters and compare its performance to that of the QAOA algorithm with random initial parameters.*

*Hypothesis 4: The QAOA algorithm can be made more efficient by using adaptive layer strategies.*
*Experiment 4: Implement an adaptive layer strategy for the QAOA algorithm, where the number of layers or the structure of the ansatz is adjusted during the optimization process. Compare the performance of the adaptive layer strategy to fixed-layer approaches in terms of solution quality, convergence rate, and computational resources.*

*Hypothesis 5: The QAOA algorithm can be combined with classical heuristics to improve its performance.*
*Experiment 5: Develop a hybrid algorithm that combines the QAOA algorithm with classical heuristics (e.g., local search, simulated annealing) to solve combinatorial optimization problems. Compare the performance of the hybrid algorithm to that of the standalone QAOA algorithm and the classical heuristic in terms of solution quality and computational resources.*

## E. Light-induced phase transitions

In this section, we can probe the ability of the hypothesis machine to properly reference data.

*Q: Can we have a phase transition by using light?*

*GPT-4: Yes, it is possible to induce a phase transition using light. This phenomenon is known as photo-induced phase transition. A phase transition occurs when a material undergoes a change in its physical properties due to a change in temperature, pressure, or other external factors. In photo-induced phase transitions, light is used as the external stimulus to induce the transition.*

*There are various ways in which light can induce a phase transition, depending on the material and the properties that are being changed. For example, light can be used to change the temperature of a material, which can cause a phase transition if the material is near its critical temperature. Alternatively, light can be used to excite electrons in a material, which can cause changes in its electronic properties and lead to a phase transition.*

*One example of a photo-induced phase transition is the photo-induced insulator-to-metal transition in certain organic materials. When these materials are exposed to intense light, the electrons in the material are excited and can move more freely, leading to a change in their electronic properties from insulating to metallic.*

*Another example is the photo-induced melting of solids. When a solid is exposed to intense light, the energy of the light can cause the atoms or molecules in the solid to vibrate more strongly, which can lead to a breakdown of the solid structure and a transition to a liquid state.*

*==Overall, photo-induced phase transitions offer a powerful tool for manipulating the properties of materials and creating new functional materials with unique properties.==*

We highlighted this final sentence as we think it stands out in terms of the ability of the machine to objectively suggest research directions. One important goal for successful hypothesis generation is the ability to rank and sift ideas and prioritize some directions over others.

In the next step, one can prod the machine to propose a specific experiment to test light-induced phase transitions.

*Q: Can you design an experiment to test photo-induced phase transitions?*

*GPT-4: Certainly, I can suggest an experiment to test photo-induced phase transitions. Here's a simple experimental setup that you can use:*

*Materials needed:*
*A sample of material that undergoes a photo-induced phase transition*
*A light source (laser or high-intensity lamp)*
*A temperature control system (heat source and thermometer)*
*A spectrometer or other instrument to measure the material's electronic properties*

*Procedure:*
*Prepare the sample of material that undergoes a photo-induced phase transition. This may involve growing or synthesizing the material or obtaining it from a commercial source.*
*Place the sample in a temperature-controlled environment to ensure that the sample remains at a constant temperature throughout the experiment.*
*Focus the light source (laser or high-intensity lamp) onto the sample and increase the light intensity gradually.*
*Monitor the temperature of the sample using a thermometer, and record the temperature at which the phase transition occurs.*
*Measure the material's electronic properties (e.g., conductivity or absorption spectrum) before and after the phase transition using a spectrometer or other instrument.*
*Repeat the experiment several times with different light intensities and sample temperatures to obtain a full understanding of the phase transition process.*

*Note: The specific details of the experiment will depend on the material being studied and the properties being measured. The above procedure is meant as a general guideline.*

The experimental set-up suggested as a means of testing the phase transition is perfectly reasonable. However, ambiguities are apparent. No actual material is specified. The temperature reading is indeed a way of detecting a phase transition – but the relevant quantity is in fact the heat capacity.

The approach to terms which are recognized in specific subfields is varied. Below is an example of a conversation on the "opto-caloric" -- a recognized subgenre of research on light-matter interaction.

*Q: What is the optocaloric effect?*
*The optocaloric effect is a phenomenon in which the temperature of a material changes when it is exposed to light. Specifically, when a material absorbs light, the absorbed energy can be converted into heat, causing the material to heat up. However, in certain materials, the absorbed energy can also cause the material to undergo a structural change that results in cooling instead of heating. This is known as the optocaloric effect.*
*The optocaloric effect is a promising phenomenon for use in cooling technology, as it has the potential to be more efficient and environmentally friendly than current cooling technologies. Researchers are actively studying the optocaloric effect in various materials, such as certain types of crystals and polymers, in order to develop new cooling technologies based on this phenomenon.*

This succinct answer does properly define the notion of an "optocaloric effect" as recognized in the community, and it correctly highlights the promising directions of research – in cooling technologies – which would prod and motivate hypothesis formulation in those directions.

A proper formulation of hypotheses requires understanding the fundamental mechanism governing a particular effect.

*Q: How does optical refrigeration work?*
*Optical refrigeration is a cooling technique that uses light to cool down a material by transferring energy from the material to the light. The basic principle behind optical refrigeration is the optocaloric effect, which describes how the temperature of a material can be changed by the absorption or emission of light.*
*When a material that is capable of undergoing the optocaloric effect is exposed to laser light of a specific wavelength, the material absorbs the light and undergoes a structural change. This structural change can cause the material to release heat and cool down. The key to optical refrigeration is to choose a material that can efficiently absorb and emit light at the desired wavelengths. One such material is ytterbium-doped glass, which is commonly used in optical refrigeration experiments. When ytterbium-doped glass absorbs laser light, it undergoes a structural change that causes it to release heat and cool down.*
*To achieve cooling through optical refrigeration, a laser is typically used to illuminate the material being cooled. The laser light is focused on the material, and the material absorbs the light and undergoes a structural change. This structural change causes the material to release heat and cool down.*
*Optical refrigeration has the potential to be a more efficient and environmentally friendly cooling technology than traditional cooling methods, such as refrigeration by compressing and expanding gases. However, it is currently limited to cooling small volumes of materials to very low temperatures, and it is not yet widely used for practical applications.*

This response contains the ingredients necessary to properly understand the underlying principle behind optical refrigeration.
When prompted for a specific material platform,
*Q: What makes a good material for optical refrigeration?*

The result is a rather vague discussion.
*A good material for optical refrigeration is one that exhibits a strong and reversible optocaloric effect, meaning that it can absorb light at specific wavelengths and undergo a structural change that causes it to release heat and cool down, and then return to its original state when the light is turned off.*

*In addition to its optocaloric properties, a good material for optical refrigeration should have a high optical quality, which means it should be able to efficiently absorb and emit light at the desired wavelengths. The material should also have low thermal conductivity to prevent heat from being conducted away from the cooled region.*

*The material should also be compatible with the laser used for illumination, and be able to withstand the high laser intensities required for cooling. The material should also be stable and not degrade or decompose over time under the laser illumination.*

*Finally, a good material for optical refrigeration should be efficient, meaning it should be able to produce a large cooling effect with a minimum input of energy.*

*Overall, the choice of material for optical refrigeration depends on various factors, including the specific application, the desired cooling temperature range, and the availability and cost of the material. Researchers are actively studying a wide range of materials, including metal-organic frameworks, glasses, semiconductors, and other materials, to identify the most promising candidates for optical refrigeration.*

When proctored with specific materials (from a previous literary survey on the topic) one can generate valid scientific hypotheses,

*Q: Is LiYF4 good for optical refrigeration?*

*Yes, LiYF4 (Lithium Yttrium Fluoride) is a promising material for optical refrigeration. LiYF4 exhibits a strong and reversible optocaloric effect, which makes it suitable for optical refrigeration applications. It has a relatively low thermal conductivity, which means that it can be efficiently cooled using light.*

*Experimental studies have shown that LiYF4 can be cooled to temperatures as low as 155 K using laser cooling techniques. Researchers are continuing to investigate the potential of LiYF4 for optical refrigeration, and it is considered one of the most promising materials for this application.*

*Overall, while LiYF4 has some desirable properties for optical refrigeration, further research is needed to optimize the cooling efficiency and to develop practical applications of this technology.*

In this case, one test the replacement of the metallic cation with a heavier element,

*Q: What if Li is replaced by K in this compound?*

*If Li is replaced by K in LiYF4 to form KYF4 (Potassium Yttrium Fluoride), the resulting compound would have similar crystal structure and similar properties, but with some differences.*

*One of the main differences between LiYF4 and KYF4 is the atomic mass of the cation. Since potassium (K) has a larger atomic mass than lithium (Li), KYF4 is expected to have a lower Debye temperature and a higher thermal conductivity compared to LiYF4. This could potentially affect the optocaloric effect and the efficiency of optical refrigeration.*

*However, KYF4 is also expected to have a stronger anharmonicity in its lattice dynamics, which could enhance its optocaloric effect and make it more suitable for optical refrigeration applications.*

*There have been some experimental studies on the optocaloric effect of KYF4, and it has been shown to exhibit a strong and reversible optocaloric effect similar to LiYF4. However, more research is needed to fully understand the optical refrigeration potential of KYF4 and to compare its performance with LiYF4 and other candidate materials.*

The conjectured presented can be verified by first-principle calculation or indeed by a direct experiment. The deductive reasoning of GPT4, based on the general trends observed when elements are replaced by heavier counterparts is an important facet of its ability to generate valid hypotheses.

# F. Data Sources and Computing Power

*1: Where does your dataset come from?*
*GPT-4: As an AI language model, my dataset comes from a diverse range of sources, which were compiled by OpenAI during the training process. The data used to train me, GPT-4, was collected up until September 2021. The dataset includes sources such as:*
1. *Books, articles, and academic papers from various disciplines.*
2. *Websites, blogs, and online forums, covering a wide array of topics.*
3. *News articles from reputable sources across different regions and time periods.*
4. *Social media content, including conversations and posts from platforms like Twitter and Reddit.*
5. *Conversational data from AI trainers, who interact with the model to help improve its understanding and responses.*

*The dataset is filtered and processed to remove low-quality data and to ensure a diverse and representative sample of text. However, it's important to note that the dataset may still contain biases and inaccuracies, as it reflects the information available on the internet up until September 2021.*

*2: Do they include academic journals behind paywalls, and back to when?*
*GPT-4: The dataset used to train me includes content from academic journals, including some that might be behind paywalls. However, the extent to which paywalled content is included in the training data is limited, as it primarily relies on publicly available information on the internet up until September 2021.*

*The time frame for the inclusion of academic journals in the dataset can vary. In general, the dataset contains content from several decades, with some sources dating back to the early 20th century or even earlier. But the representation of paywalled journals in the dataset might not be as comprehensive as that of publicly available sources.*

*It's important to note that my knowledge is not exhaustive, and my familiarity with specific academic journals or their contents might be limited.*

*3: How many tunable parameters in your neural network?*
*GPT-4: As a hypothetical model based on GPT-4, my exact number of parameters is not specified. However, GPT-3, which is one of the previous iterations, has 175 billion parameters. Considering that each new version of GPT tends to increase the number of parameters, GPT-4 would likely have more than 175 billion parameters.*

*These parameters are the weights and biases in the neural network that are adjusted during the training process to help the model learn patterns and relationships within the data. Having a large number of parameters allows the model to capture more complex patterns and generate more accurate and coherent responses. However, it also increases the computational requirements for training and using the model.*